\documentclass[journal,twoside,web]{ieeecolor}

\usepackage{tmi}
\usepackage{cite}

\usepackage{multirow,array,hhline}
\usepackage{framed,multirow}

\usepackage{graphicx}
\usepackage{babel}
\usepackage{amsmath}

\usepackage{subcaption}
\usepackage{amssymb}
\usepackage{latexsym}
\usepackage{url}
\usepackage{xargs}
\usepackage{textcomp}

\def\BibTeX{{\rm B\kern-.05em{\sc i\kern-.025em b}\kern-.08em
		T\kern-.1667em\lower.7ex\hbox{E}\kern-.125emX}}
\begin{document}
	\bstctlcite{IEEEexample:BSTcontrol}
	
	\title{Building a Synthetic Vascular Model: Evaluation in an Intracranial Aneurysms Detection Scenario}

\author{Rafic Nader, Florent Autrusseau, Vincent L'Allinec and Romain Bourcier
	\thanks{This work was supported by the French RHU-ANR project “eCAN” \# ANR-23-RHUS-0013 and INSERM CoPoC \# MAT-PI-22155-A-01}
	\thanks{Authors R.N, F.A and R.B are with Nantes Université, CNRS, INSERM, l’institut du thorax, F-44000 Nantes, France. France (e-mails: Rafic.Nader@univ-nantes.fr, Florent.Autrusseau@univ-nantes.fr, Romain.Bourcier@univ-nantes.fr), R.B is also with CHU Nantes, F.A. is also with the LTeN, Polytech, rue Ch. Pauc 44306 Nantes, France; author V.L. is with CHU Angers (Vincent.LAllinec@chu-angers.fr). 
		Authors R.N. and F.A. contributed equally.}}

\maketitle

\begin{abstract}
	We hereby present a full synthetic model, able to mimic the various constituents of the cerebral vascular tree, including the cerebral arteries, bifurcations and intracranial aneurysms. This model intends to provide a substantial dataset of brain arteries which could be used by a 3D convolutional neural network to efficiently detect Intra-Cranial Aneurysms. The cerebral aneurysms most often occur on a particular structure of the vascular tree named the Circle of Willis. Various studies have been conducted to detect and monitor the aneurysms and those based on Deep Learning achieve the best performance. Specifically, in this work, we propose a full synthetic 3D model able to mimic the brain vasculature as acquired by Magnetic Resonance Angiography, Time Of Flight principle. Among the various MRI modalities, this latter allows for a good rendering of the blood vessels and is non-invasive.
	Our model has been designed to simultaneously mimic the arteries' geometry, the aneurysm shape, and the background noise. The vascular tree geometry is modeled thanks to an interpolation with 3D Spline functions, and the statistical properties of the background noise is collected from angiography acquisitions and reproduced within the model. In this work, we thoroughly describe the synthetic vasculature model, we build up a neural network designed for aneurysm segmentation and detection,  finally, we carry out an in-depth evaluation of the performance gap gained thanks to the synthetic model data augmentation.
\end{abstract}


\section{Introduction}

\IEEEPARstart{T}{his} work has been carried out in the context of a wide medical research project in which neuroradiologists intend to estimate the risk of occurrence and/or rupture of Intra-Cranial Aneurysms (ICA)~\cite{Bourcier2017}. 
The advent of ICA formation results from various factors, among which the genetic risk seems predominant \cite{Zhou2018}. 
However, it is commonly accepted among physicians that the geometric disposition of the cerebral vascular tree might explain why a weakened vessel wall (due to genetic or environmental factors) might give rise to an aneurysm.
Untreated brain aneurysms pose a significant risk of rupture, which can result in a hemorrhagic stroke. In fact, this rupture can potentially lead to the patient's death in as much as $50\%$ of all cases.
Magnetic Resonance Angiography (MRA), Time of Flight (TOF) modality is frequently used for aneurysms detection \cite{sailer2014}. Unlike other methods like Digital Subtraction Angiography (DSA) and Computed Tomographic Angiography (CTA), TOF is radiation-free and doesn't require the administration of a contrast agent~\cite{Summerlin2022}. Given the mounting workload and the demanding nature of the detection process undertaken by radiologists, there is an increasing  need for an automated tool to detect and monitor aneurysms at an early stage.
Prior to the widespread adoption of Deep Learning (DL), research studies employed imaging filters or traditional machine learning techniques to detect aneurysms \cite{zeng,yang2011computer}. 
Recent advances in artificial intelligence, particularly those involving deep Convolutional Neural Networks (CNNs), have significantly enhanced the development of automatic tools in the field of medical imaging \cite{wang2012machine}.
To date, several deep learning based approaches have been proposed for ICA segmentation and/or detection \cite{stember2019,chen2020,joo2020,Timmins2021}. 
The ADAM Challenge compared 11 different DL approaches for detecting or/and segmenting ICAs on TOFs.
The best algorithm, \cite{baumgartner2021} achieved notable results, with an average sensitivity of $0.67$ and a false positive rate of $0.13$.
It is important to note that a majority of the existing methods have been formulated using private clinical data that comes with meticulously refined manual annotations. Indeed, one of the obstacles in developing deep learning methods for medical imaging applications is the lack of large annotated datasets, particularly for the segmentation task. To mitigate this, Di Noto et al. \cite{di2023} proposed  the use of ``weak'' annotations and they obtained good results with an average lesion sensitivity of $0.83$ and a false positive rate of $0.8$. Other studies adopt
non-voxel-based methods, such as mesh convolutional neural
networks \cite{timmins2023}, to overcome limitations related to modality and
scan acquisition parameters.

The rationale behind our work is to try to reduce as much as possible, or even to free oneself from any manual labeling. In other words, we expect that using several hundreds or even thousands of modeled bifurcation to train a network might provide better performances than using only  a couple of hundreds actual TOF segmentations.
Unlike previous works, in our approach, we investigate the brain aneurysm detection task by exploiting synthetic data.
While data augmentation stands out as a well-known technique for augmenting the number of data samples, its application requires careful consideration.
In the context of medical images, such image manipulations might tamper with the geometrical or statistical properties in an undesirable way, \textit{i.e.} render the augmented images too distant from their corresponding ground truth.

In the past, several works have been devoted to the design of computer models intended to mimic arterial trees. At that time, the studies focused on constrained constructive optimization~\cite{Karch1999}. Some models were particularly designed to offer a high graphical fidelity through a better understanding of the biophysical properties~\cite{Szekely2002}. A relatively nice rendering was obtained on liver vascular trees for instance in~\cite{Coatrieux2003}.
Such models were mostly designed in the aim to study angiogenesis (physiological process leading to the formation of new blood vessels.).
More recently, the VascuSynth model~\cite{Hamarneh2010} was proposed in the aim to produce vast amounts of volumetric vasculature images. Here, the aim was different, the authors intended to generate a synthetic  dataset for image segmentation. 
All these computer models achieved a quite accurate modeling of the acquired medical images (mostly trying to mimic liver or lung vasculatures).
However, modeling the cerebral vascular tree is more challenging, as the arteries are commonly longer, and may exhibit a stronger tortuosity.
Moreover, in our study, the goal strongly differs. We intend to generate vast amounts of images to train a neural network for a pattern recognition task.
Another interesting approach was proposed in ~\cite{subramaniam2022generating,kossen2021synthesizing} where the authors generate synthetic MRI patches using Generative Adversarial Networks. Indeed, random patches were generated along with their underlying ground truth segmentation. 
This method has not been designed to generate a given target bifurcation, or to add an aneurysm onto the vascular tree, which is an issue we tackle in our work. 

In a previous study~\cite{ICPR2022}, we have proposed a fully synthetic model of 3D bifurcations and Intra-Cranial Aneurysms. 
In this work, we tried to mimic the geometrical shape of arterial bifurcations by generating linear segments to be later convolved by a spherical kernel, and apply some geometric distortions to model the tortuosity. 
A particular attention was devoted to the generation of a plausible background noise. 
Ultimately, an intracranial aneurysm was modeled and superimposed onto the bifurcation. 
Although this initial model proved to mimic relatively well 3D bifurcations, it showed some limitations, regarding the tortuosity, and the aneurysm location (slightly shifted away from the bifurcation). 
In the current work, our model accuracy is considerably increased, and we intend to propose a set of highly realistic bifurcations and aneurysms.
Since our intended application involves identifying an aneurysm within a specific bifurcation or artery from a TOF scan, it is crucial to accurately model various essential arterial features: the shape, orientation, diameters, and tortuosity. As for the aneurysms, the model should allow to adjoin ICAs of various shapes and sizes onto different bifurcations. 
Finally, the performances in terms of image segmentation might depend on the accuracy of the modeled surrounding background noise, hence, it is important for the model to faithfully duplicate the background noise.

In section~\ref{sec:MnM}, we thoroughly describe the synthetic model. Its three main features are presented, namely \textit{i)} the arteries geometry, \textit{ii)} the surrounding TOF noise and \textit{iii)} the modeled aneurysm.
Next, in section~\ref{sec:ExpRes} we provide an in-depth description of the generated dataset, we describe the CNN architecture, evaluate both the ICA segmentation and detection performances. We try to assess the performance gain brought by using the synthetic images alongside the manually labeled ones. To do so, we run two separate experiments involving either the manually segmented images only or adjoining the modeled patches.
Finally, in section~\ref{sec:Discussion}, we discuss the benefit of using the synthetic model for intracranial aneurysm detection and conclude this work.

\section{Material and Method}
\label{sec:MnM}

\begin{figure}[!ht]
\begin{centering}
\includegraphics[width=0.7\columnwidth]{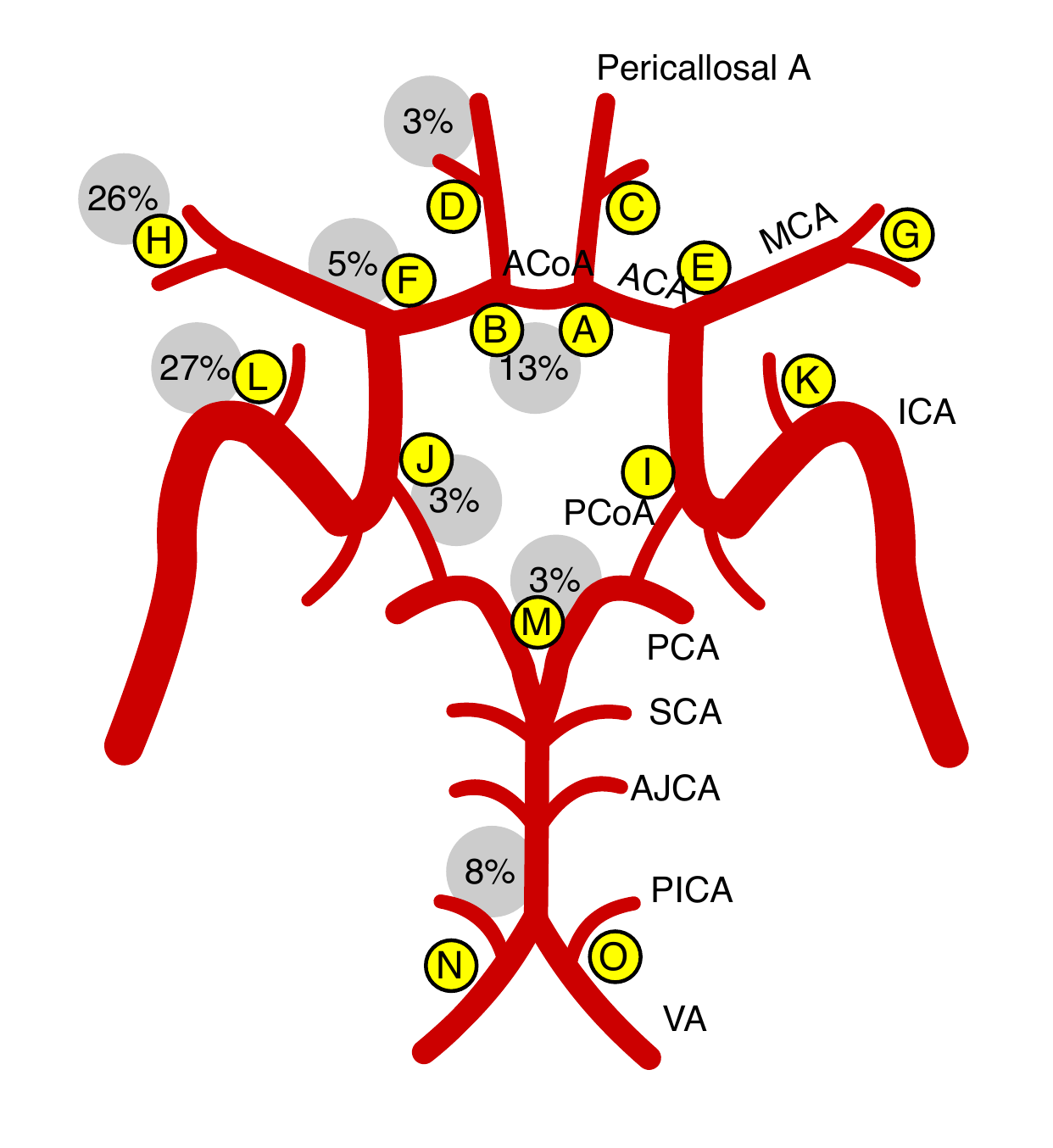}
\par\end{centering}
\caption{Schematic representation of the CoW. The yellow labels (from A to O) depict the particular bifurcations we are interested in for this study. The percentages within the gray discs represent the risk of aneurysm formation. \label{fig:SchematicWillis}}
\end{figure}

Unlike the previous model, proposed in \cite{ICPR2022}, here we intend to come up with a full synthetic model of 3D cropped TOF portions, including not only the bifurcation of interest, but also its whole neighborhood\footnote{Source code for the synthetic Vascular Models (VaMos) available here:\\ \url{https://gitlab.univ-nantes.fr/autrusseau-f/vamos/}}.
Hence, we propose a completely new method here. Within a 3D cropped portion of a MRA volume, we collect the 3D coordinates of the arteries' skeleton (centerline of the 3D tubes), and we further fit those centerlines using 3D splines.
It is quite vastly admitted that there can be a significant structural variability in the vasculature of individuals \cite{Robben2016}. Indeed, one can find relatively strong variations among the shape of the Circle of Willis (CoW) (some arteries and bifurcations may be missing for some patients, or even some extra bifurcations could be present for others), and the anatomical properties of the cerebral arteries can also strongly differ from one person to another. The arteries' angles, their tortuosity, diameters or even their geodesic length can significantly differ.
A schematic representation of the CoW is given in Fig. \ref{fig:SchematicWillis}. 

\begin{figure*}[!ht]
\begin{centering}
\includegraphics[width=\textwidth]{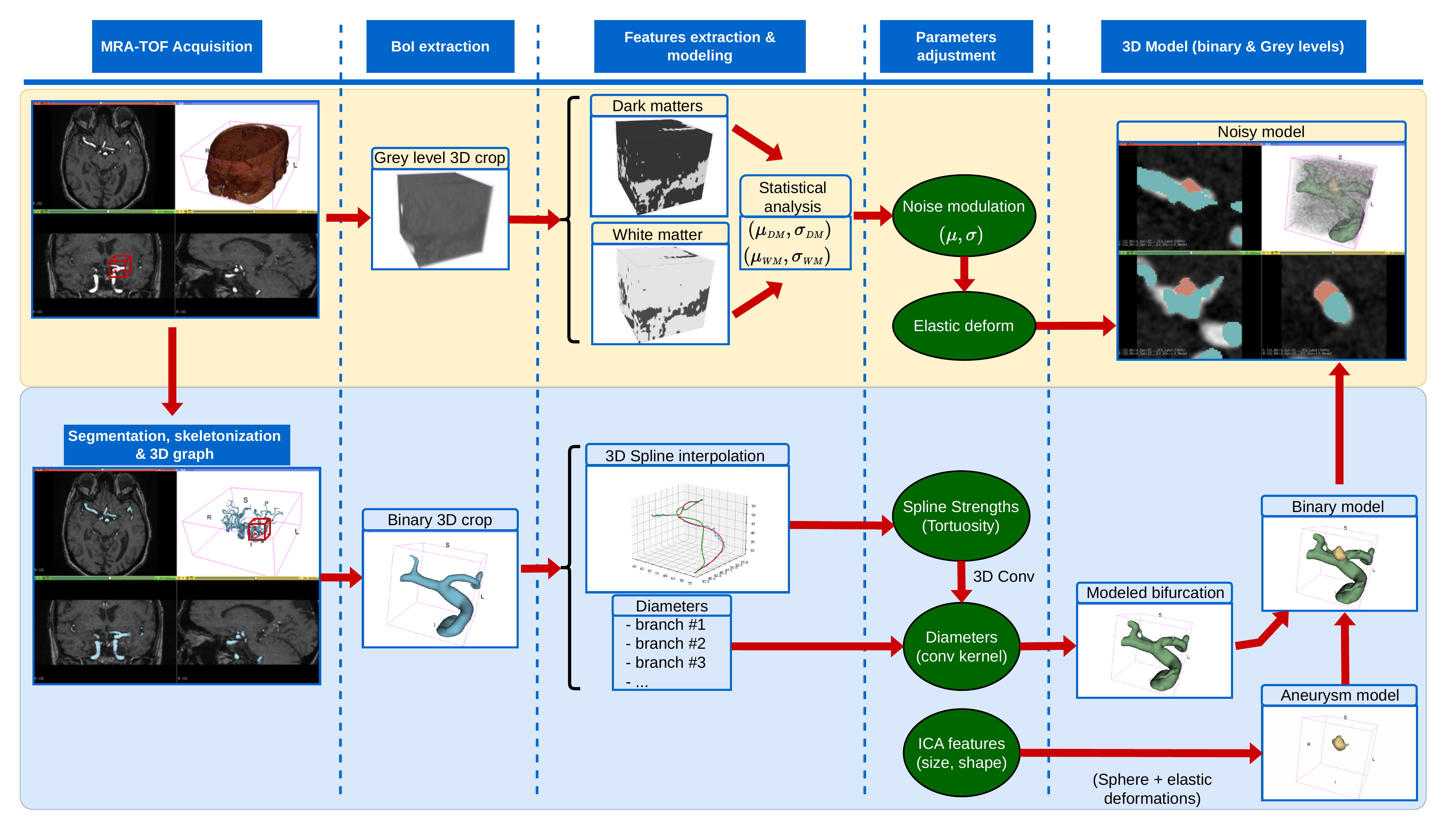}
\par\end{centering}
\caption{Schematic representation of the whole bifurcation model. The upper part (yellow
shaded block) represents the background noise modeling, whereas the lower part
(light blue shaded block), shows the arterial geometry modeling. The green ellipses represent the different features the synthetic model can modify.\label{fig:VaMosProcess}}
\end{figure*}

Such an important variability can make the task quite difficult for neural networks to properly recognize and/or segment the bifurcations/arteries of interest.
That is why our aim here is to propose a model being able to generate a vast amount of highly similar arteries, while adjusting at will the geometric features and the grey level amplitude of the vasculature, or even the statistical properties of the surrounding noise.
Not only does our model consider the blood vessels geometry, but it also able to accurately counterfeit the background noise.
The full process of the vasculature model is represented in Fig. \ref{fig:VaMosProcess}. The blue rectangles represent the different steps needed to produce one bifurcation model, the green ellipses show the various parameters we can modulate to distort the ICA, the bifurcation or the background.

Let us first focus on the shape modeling. 
The goal being to come up with similar shapes that can be slightly modified, but with the strong requirement to still best represent actual human arteries, the mimicking of true TOF components is crucial.

\subsection{Modeling the arterial geometry}
\label{ssec:ModGeomArt}

As a very first step, we extract a 3D graph from the segmented TOF volume (interested reader may refer to~\cite{Nouri2020} for further details on the graph method being used). Obviously, any other 3D graph method, such as the one presented in~\cite{Damseh2021} could be used.
Such a graph is simply composed of 3D branches and nodes. A node is a branch extremity, \textit{i.e.} either a bifurcation or a branch end point. From a binary segmented vasculature, the 3D graph allows to locate a given bifurcation among the ones composing the CoW and to extract a whole 3D crop around the 3D coordinates of the bifurcation node.
The bifurcations of interest are automatically located thanks to the works from~\cite{NADER2023102919}. A 3D patch is cropped around the $(x,y,z)$ coordinates of the detected bifurcation of interest (\textit{i.e.} bifurcation forming the CoW). 
Within this cropped portion of the segmented TOF, the set of all coordinates along the branches are collected and curve fitting is used to represent the points using 3D splines functions \cite{Splines93}.

Spline functions can be represented by three different features:
\textit{i)} the knot-points, defining the intervals of the chunks on which the polynomials are defined, \textit{ii)} the B-Splines' (or polynomials') coefficients, and \textit{iii)} the order of the spline, (\textit{i.e.} the degree to which the fit was performed). Once these parameters have been collected for each 3D branch composing the bifurcation, they can thus be easily modified in order to distort the centerline coordinates. Specifically, in this model, we only alter the polynomials coefficients.

Once the centerlines have been tweaked via the spline function alteration, we shall collect the diameters of all arteries being accounted for within the 3D crop.
This can easily be handled by using our previous vascular tree characterization tool \cite{Nouri2020}.
Each centerline (morphological skeleton) being first tweaked by the spline alteration, can thus go through a convolution with a spherical kernel which size is adapted to the corresponding observed diameter. We then thicken each artery according to its anatomical property.

\begin{figure}[!ht]
\centering
\subfloat[Spline model with weak weights]{ 
\includegraphics[width=0.45\columnwidth]{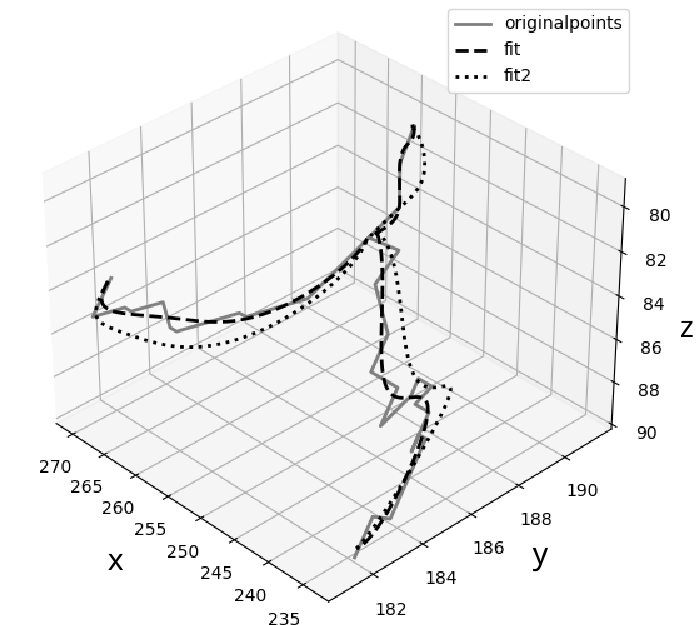}}
\hfill{}\subfloat[Spline model with larger modifications]{ 
\includegraphics[width=0.45\columnwidth]{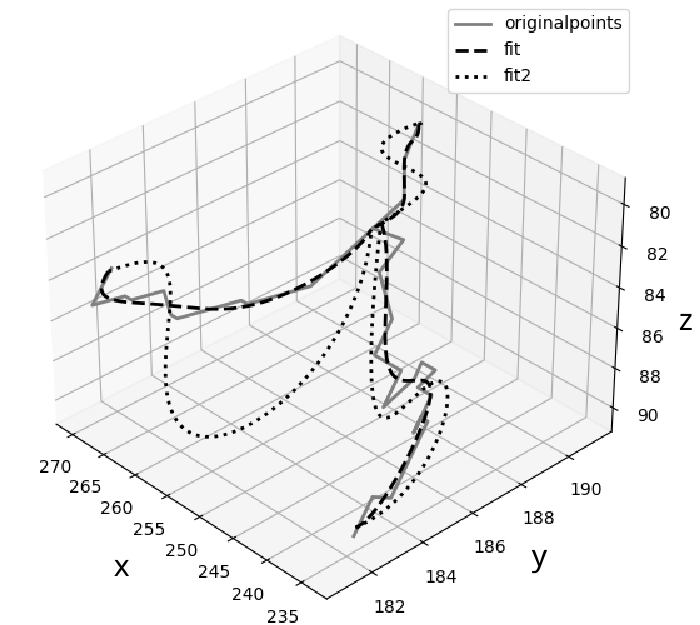}}
\caption{Examples of modified bifurcation centerlines. The solid gray line represents the actual 3D branch, the black dashed line represents the best spline fit, and the dotted line represents an exaggerated modification of the spline coefficients. \label{fig:Spline_3D_bifurc}} 
\end{figure}

We show on Fig. \ref{fig:Spline_3D_bifurc} some examples of 3D Spline models. The plots show, for a given bifurcation, three different 3D representations. The solid gray lines represent the actual bifurcation coordinates, as collected within the TOF acquisition, the black dashed lines stand for the spline functions that best fit the arteries, and finally, the black dotted curves show the altered spline function (the new centerline of the bifurcation to be).
Besides the convolution with a spherical kernel, that is needed to set the artery's thickness, our method allows to set a target grey level amplitude. We simply multiply the binary envelope of the modeled vasculature by the desired target grey level. Moreover, for the thickened artery to have a realistic shape, it is important that the convolution kernel is not perfectly spherical. 
We hence apply elastic deformations to avoid producing a perfectly tubular modeled artery.

\subsection{Modeling and adjoining the aneurysm}
\label{ssec:ICApos}

\begin{figure}[!ht]
\begin{centering}
\includegraphics[width=0.85\columnwidth]{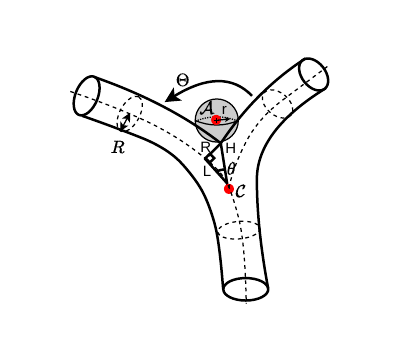}
\par\end{centering}
\caption{Computation of the distance separating the aneurysm and the bifurcation center. The ICA is located along the bisector, the distance separating points $\mathcal{A}$ (ICA center) and $\mathcal{C}$ (bifurcation node) needs to be estimated (see text).\label{fig:ICAPos}}
\end{figure}

Once the bifurcation has properly been modeled, a synthetic aneurysm can finally be incorporated.
A simple 3D sphere is first created and then distorted using elastic deformations. 
We use the geometric distortions from~\cite{Ronneberger2015} (with $\sigma\in[0,6]$ and a $3\times3$ deformation grid). 
Further, the ICA center is aligned onto the bisector between the two daughter arteries. 
Figure~\ref{fig:ICAPos} shows a configuration of aneurysm positioning.

The distance between point $\mathcal{C}$ and $\mathcal{A}$ being $H+r$, we can easily compute H, as follows:

\begin{equation}
\tan{\theta} = \frac{R}{L}
\end{equation}

\begin{equation}
L=\sqrt{H^2-R^2}
\end{equation}

Hence, 

\begin{equation}
\sqrt{H^2-R^2} = \frac{R}{\tan{\theta}}
\end{equation}

and, thus:

\begin{equation}
H=\sqrt{\left( \frac{R}{\tan{\theta}} \right)^2 + R^2}
\end{equation}

Then, considering a growth parameter ($\gamma$), being applied to modulate the distance $H$, the distance from the aneurysm center to the bifurcation node has been computed as in Eq.~\ref{eq:Distance_ICA} below.
\begin{equation}
\mathcal{D} = r \times \gamma +\sqrt{\left(\frac{R}{tan(\Theta/2)}\right)^2+R^2}
\label{eq:Distance_ICA}
\end{equation}
where $r$ is the ICA radius, $R$ is the radius of the bifurcation branches , and $\Theta$ stands for the angle formed by the two daughter arteries (we assume $\theta=\Theta/2$). Thanks to the growth parameter $(\gamma)$, we can automatically adjust the shift from the aneurysm center and the vessel wall where the daughter arteries split. Hence, we can model various states of growth for a given aneurysm (\textit{i.e.} a neck being more or less present).

In the forthcoming experiments, the various parameters of the synthetic model were chosen, in such a way to best represent the anatomical properties of the vasculature, \textit{i.e.} we have conducted a thorough
evaluation of both the modeled bifurcations and the aneurysm sacs. Concerning the bifurcations, we ensured that the order of magnitude of the angles, diameters and tortuosities were respected by the model. 
As for the aneurysms, various geometric features were compared, such as the volume, surface, sphericity, elongation and flatness. 
This analysis led us to determine the model parameters as follows :
The ICAs radii were chosen such that $r\in[0.4, 2.0]$ mm, the ICA growth parameter was in the range $\gamma\in[0.7, 1.0]$, the spherical ICA was deformed with $\sigma\in[1.0, 4.0]$, and the B-Spline coefficients was set to $2$.

\subsection{Modeling the surrounding brain matters}

Now that we have been through the details of the shape modeling, let us next present the second part of the synthetic model: the brain, composed of fluids and white/gray matters, all these, being affected by a reconstruction background noise.

\subsubsection{Collecting the target statistics }

As can be observed on Fig.~\ref{fig:VaMosProcess}, for the background generation (brain and noise), the cropped TOF patch first goes through a separation of the various brain components.
In fact, based on the same rationale that was previously used (section~\ref{ssec:ModGeomArt}) to model the arteries, we can easily imagine to include the various brain matters within the model. Indeed, the cerebral arteries are surrounded by various matters, each one having a particular radio-opacity, \textsl{i.e.} a different gray level. The white/gray matter, when acquired through MRA appears with relatively high gray levels, the Cerebro-Spinal Fluids (CSF), the ventricle or the Corpus Callosum are commonly displayed with much lower luminances.
However, unlike our previous study \cite{ICPR2022}, the fluid areas (hypointense signal) within the brain are no more randomly determined; we believe that those local low contrast shapes may be of paramount importance while modeling the 3D crops of our arteries and bifurcation, and hence we intend to include a faithful representation of  the fluids areas within the synthetic model.
The separation between hypointense and hyperintense matters (vasculature excluded) were obtained via a simple multi-threshold segmentation~\cite{Liao2001}. 
Once located, each matter can then be geometrically distorted, before generating and applying it's overlaying (modified) noise.

\subsubsection{Noise generation}
\label{sssec:Noise}

When a Gaussian blur is being applied onto an input image $I(x,y)$ its filtered version becomes: 

\begin{equation}
O(x,y)=\sum_{i=-\infty}^{\infty}\sum_{j=-\infty}^{\infty}\frac{1}{2\pi\sigma_{G}^{2}}e^{-\frac{i^{2}+j^{2}}{2\sigma_{G}^{2}}}I(x+i,y+j)\label{eq:GBlur}
\end{equation}

The Bienaym\'e's identity states that

\begin{equation}
\begin{array}{c}
Var\left(\sum_{i=1}^{n}X_{i}\right)=\\
\sum_{i=1}^{n}Var(X_{i})+\sum_{i,j=1,i\neq j}^{n}Cov(X_{i},X_{j})
\end{array}\label{eq:Bienayme}
\end{equation}

Thus, the variance of a linear combination is:

\begin{equation}
\begin{array}{c}
Var\left(\sum_{i=1}^{n}c_{i}X_{i}\right)=\\
\sum_{i=1}^{n}c_{i}^{2}Var(X_{i})+2\times\sum_{i,j=1,i\neq j}^{n}c_{i}c_{j}Cov(X_{i},X_{j})
\end{array}\label{eq:VarLinearCombination}
\end{equation}

However, if $X_{i},...,X_{n}$ are pairwise independent integrable random variables $\left(Cov(X_{i},X_{j})=0,\,\forall(i\ne j)\right)$, which we assume in the following, then:

\begin{equation}
Var\left(\sum_{i}c_{i}X_{i}\right)=\sum_{i}c_{i}^{2}Var(X_{i})\label{eq:VarIdentity2}
\end{equation}

where $c_{i}$ are constants.
We consider that the variance of the input image is $Var\left[I(x+i,y+i)\right]=\sigma_{0}^{2}$, our goal here is to estimate the variance of the output (filtered) image $Var\left[O(x,y)\right]=\sigma_{f}^{2}$. Thus,

\begin{equation}
\sigma_{f}^{2}=\sigma_{0}^{2}\sum_{j=-\infty}^{\infty}\sum_{i=-\infty}^{\infty}\left(\frac{1}{2\pi\sigma_{G}^{2}}e^{-\frac{i^{2}+j^{2}}{2\sigma_{G}^{2}}}\right)^{2}\label{eq:sigma}
\end{equation}

For large $\sigma_{G}$, the squared Gaussian is smooth and the sum can be approximated as:

\begin{equation}
\begin{array}{c}
\sigma_{f}^{2}\approx\sigma_{0}^{2}\int_{-\infty}^{\infty}\int_{-\infty}^{\infty}\left(\frac{1}{2\pi\sigma_{G}^{2}}e^{-\frac{i^{2}+j^{2}}{2\sigma_{G}^{2}}}\right)^{2}di.dj\\
=\frac{\sigma_{0}^{2}}{4\pi\sigma_{G}^{2}}
\end{array}\label{eq:sigmaf_sigma0}
\end{equation}

and thus,

\begin{equation}
\sigma_{f}\approx\frac{\sigma_{0}}{2\sigma_{G}\sqrt{\pi}}\label{eq:sigmaf_sigma0-2}
\end{equation}

In summary, when an image composed of Gaussian noise of standard deviation $\sigma_{0}$ is being filtered by a Gaussian filter of standard deviation $\sigma_{G}$, the so-obtained filtered image has a standard deviation of $\sigma_{f}$ according to the Eq.~\ref{eq:sigmaf_sigma0-2}.

However, for our particular purpose, we intend to determine which Gaussian filter (of standard deviation $\sigma_{G}$) shall be used on the input image so as to obtain a filtered image with a given target statistics ($\sigma_{f}$), and hence $\sigma_{G}\approx\sigma_{0}/(2\sigma_{f}\sqrt{\pi})$.

The process starts thus with the generation of a high frequency Gaussian Noise of average set to our target 3D crop. This noise will then be smoothed out using a Gaussian filter of standard deviation $\sigma_{G}$. The resulting image (of standard deviation $\sigma_{f}$) will thus present strong similarities with the target portion of the TOF being modeled.

The so-obtained noise is then simply added up on top the geometric modeling (bifurcation and aneurysm as explained in previous sections~\ref{ssec:ModGeomArt} and~\ref{ssec:ICApos}.)

\subsection{Model examples}

\begin{figure}[!ht]
\begin{centering}
\subfloat[Bifurcations : Ground Truth \textit{vs.} Model]{ 
\includegraphics[width=0.86\columnwidth]{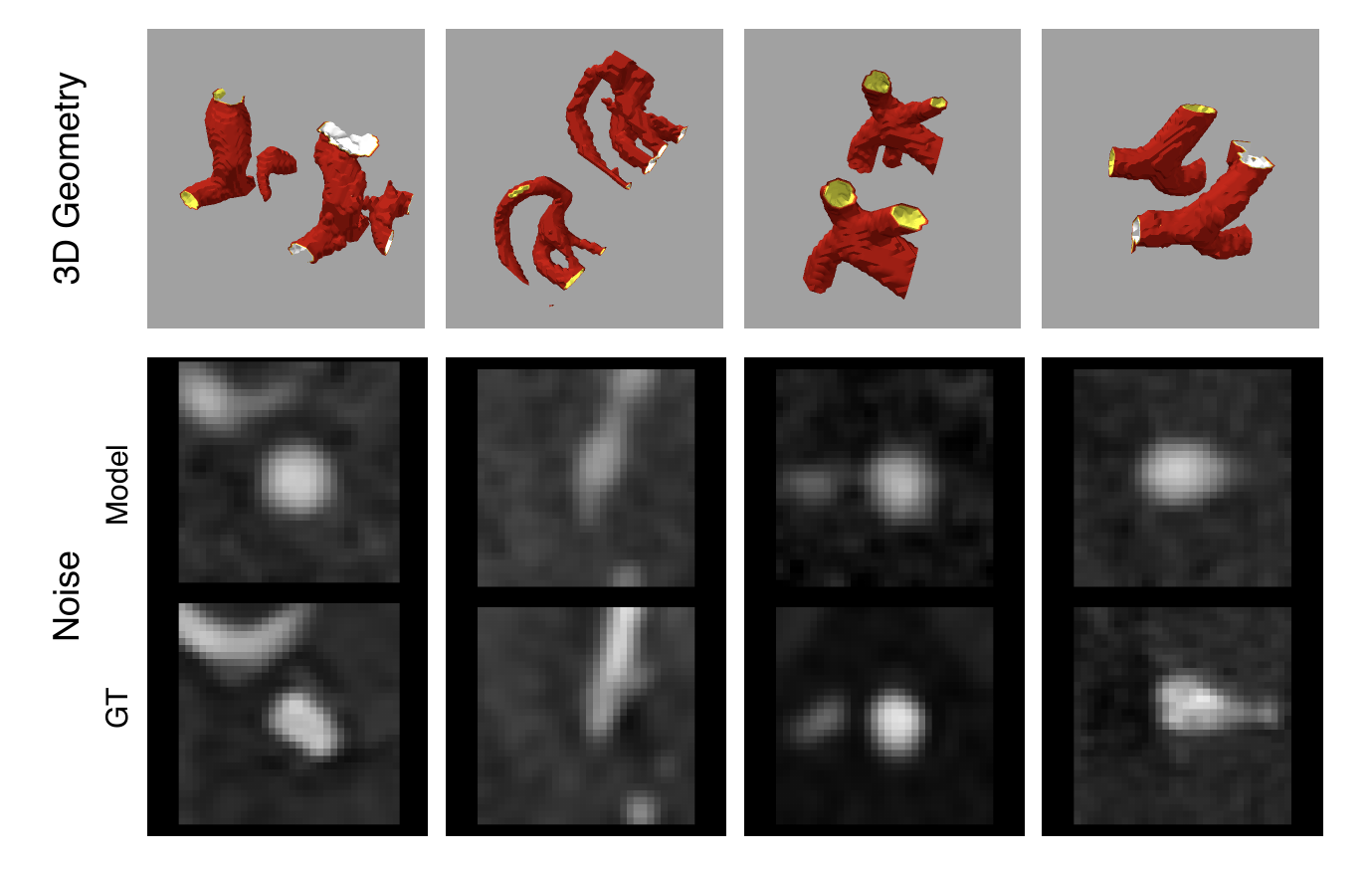}}
\hfill{}\subfloat[Aneurysms : Ground Truth \textit{vs.} Model]{ 
\includegraphics[width=0.86\columnwidth]{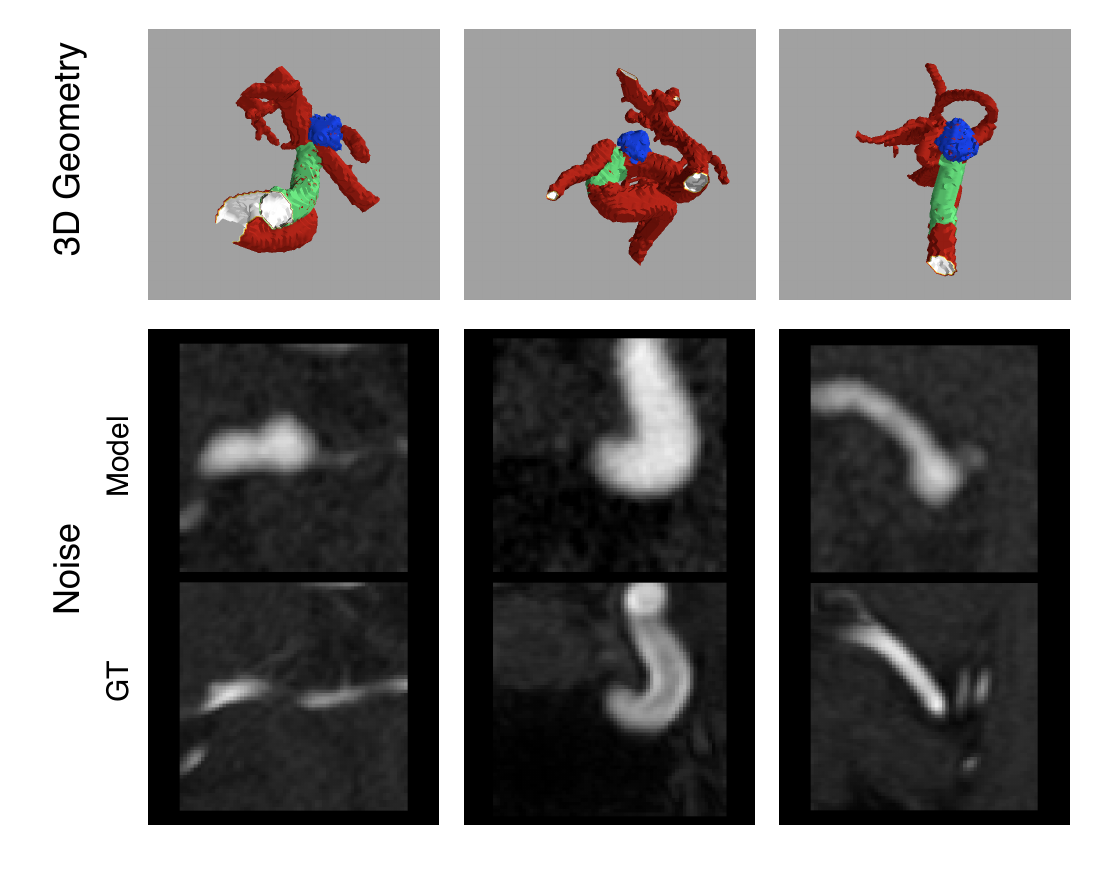}}
\end{centering}
\caption{Comparison between the modeled bifurcations and the Ground Truth crop from a TOF (with/without ICA). 
We show both the gray level voxels patches (lower panels) and 3D bifurcation layout (upper panels). \label{fig:Comp_VaMos_GT_Bif}}
\end{figure}

Concerning the bifurcation model only (\textit{i.e.} no adjoined aneurysm), Fig.~\ref{fig:Comp_VaMos_GT_Bif}(a) shows, for four different extracted bifurcation patches, a comparison between the Ground Truth (GT) crop and the modeled patch. Images on the lower panel shows a 2D slice of the gray level voxels for both the Ground Truth and the Model, whereas the upper images represent the 3D layouts of the arteries.
We can notice that both the geometrical configuration of the bifurcations and the gray level distribution seem to be efficiently modeled and mimic very accurately the TOF patch. Indeed, a very wide variety of bifurcations (no matter how complex the shapes are) can easily be modeled. We can notice that some subtle diameter or tortuosity modifications are faithfully brought on the modeled bifurcations.

Besides the bifurcations themselves, it is crucial for the aneurysm to be accurately modeled and most importantly well positioned onto the bifurcation artery wall. We present three different examples on Fig.~\ref{fig:Comp_VaMos_GT_Bif}(b). Again we can observe how the slice gray levels are faithfully represented in the model. On the upper images, the aneurysms are represented in blue, whereas the mother artery is depicted in green.

It is important to note that, so far, the synthetic aneurysm model can only reproduce the unruptured and untreated aneurysm. Indeed, ruptured or treated aneurysm presents significant differences, treated aneurysms are radiolucent (hypointense signal), and can easily be mistaken by surrounding fluids,  whereas ruptured aneurysms present very different shapes (higher order spherical aberrations, larger elongation, etc.). 

\begin{figure}[!ht]
\centering
\subfloat[PSNR]{ 
\includegraphics[width=0.45\columnwidth]{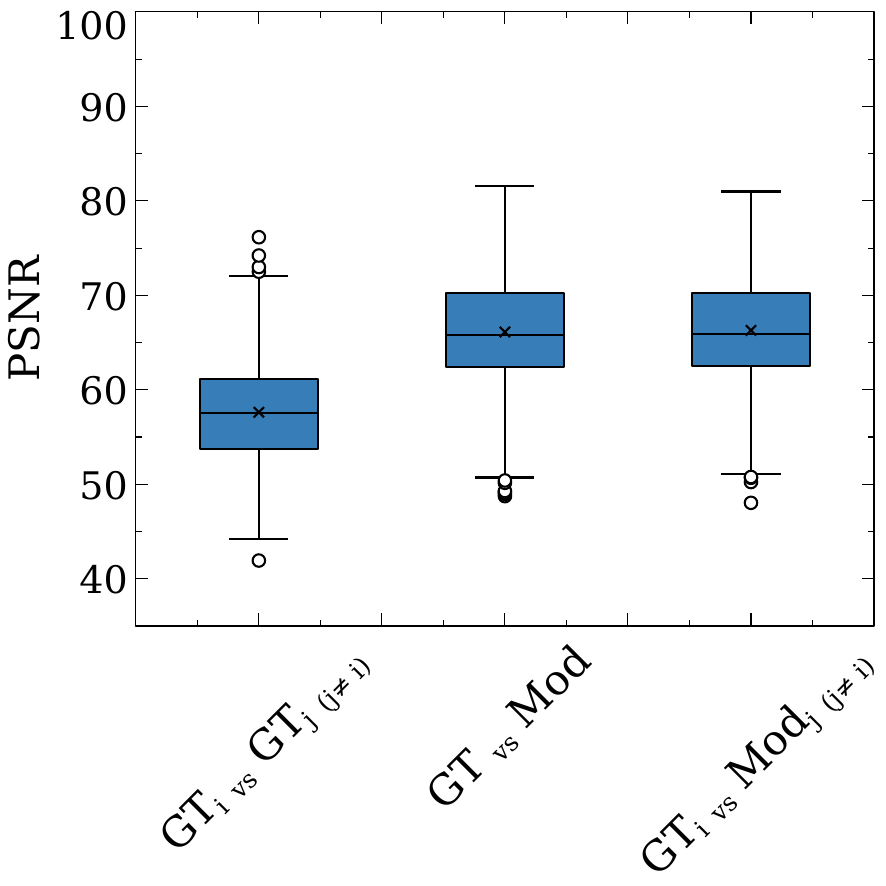}}
\hfill{}
\subfloat[VIF]{ 
\includegraphics[width=0.45\columnwidth]{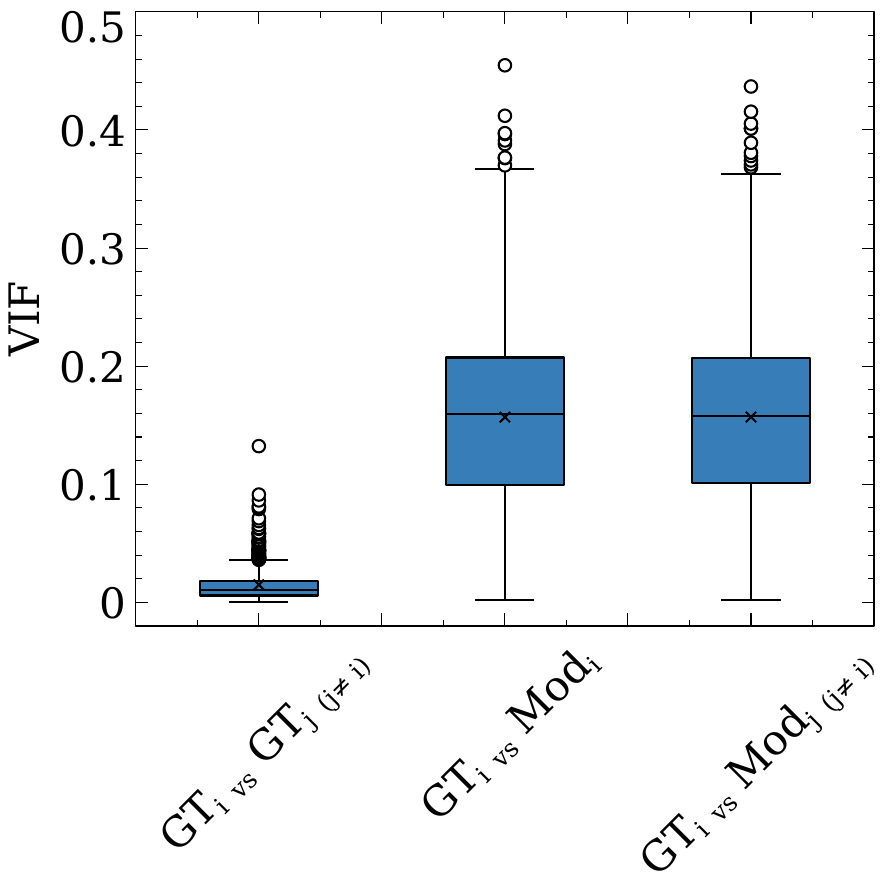}}
\caption{Objective assessment of the similarities between the ground truth patches and synthetic models.\label{fig:OQMs}}
\end{figure}

Naturally, a simple visual inspection by itself (as in Fig.~\ref{fig:OQMs}) is not sufficient to accurately estimate the faithfulness of the synthetic model. Hence, we also provide a quantitative (objective) assessment of the modeled bifurcations.

We show on Figure~\ref{fig:OQMs} the distribution of two distinct objective quality assessment metrics from $1000$ patches (2D slices randomly extracted from 150 TOF volumes). 

Overall, 9 metrics have been tested (MSE, PSNR, NRMSE, NMI, SSIM, VI, NQM, VIF and UQI)\cite{IQMs2020,NMI1999}, but most exhibited similar behaviour, and hence, we only present the results for PSNR and VIF (which was proven to perform better on MRI/CTA acquisitions~\cite{Ohashi2023, IQMs2020}).
Each sub-plot in Figure~\ref{fig:OQMs} presents, along the x-axis \textit{i)} the comparison between distinct patients, \textit{ii)} the comparison between a given ground truth patch and its modeled version, and \textit{iii)}  the comparison between a given ground truth patch and the modeled patch from a different patient. 
We can observe that overall, the modeled patches better represent a bifurcation than the patches collected from another patient. 
However, it is important to note that, providing objective quality measurements, without a ground truth validation (subjective experiment) 
is suboptimal, as it is difficult to ensure the accuracy of the tested quality metrics. Future works will be devoted to a thorough evaluation of the synthetic patches fidelity, including subjective experiments.

\subsection{Applications of such a synthetic model}
\label{ssec:ModelApplications}

Such synthetic images could be very useful for a large range of scenarios:

\begin{enumerate}
\item In a medical center having few images, one could imagine generating a massive training dataset targeting the features of the MRA acquisitions.
\item The synthetic model can be exploited for other computer vision / pattern recognition tasks, such as vasculature segmentation, bifurcations detection/classification, or even stenosis/thrombosis detection.
\item If a new MRA scanner was being brought to the market (higher magnetic field, better resolution,...), we could promptly provide a massive annotated dataset, without resorting to tedious manual annotations from expert.
\item The model can provide a relatively easy adaptation to other imaging devices (such as DSA or CTA).
\item Ultimately, such a model could be used to detect aneurysms on a healthy dataset (where no patient actually bears an aneurysm).
\end{enumerate}

\section{Experimental Results}
\label{sec:ExpRes}

\begin{figure*}[!ht]
\begin{centering}
\includegraphics[width=0.92\textwidth]{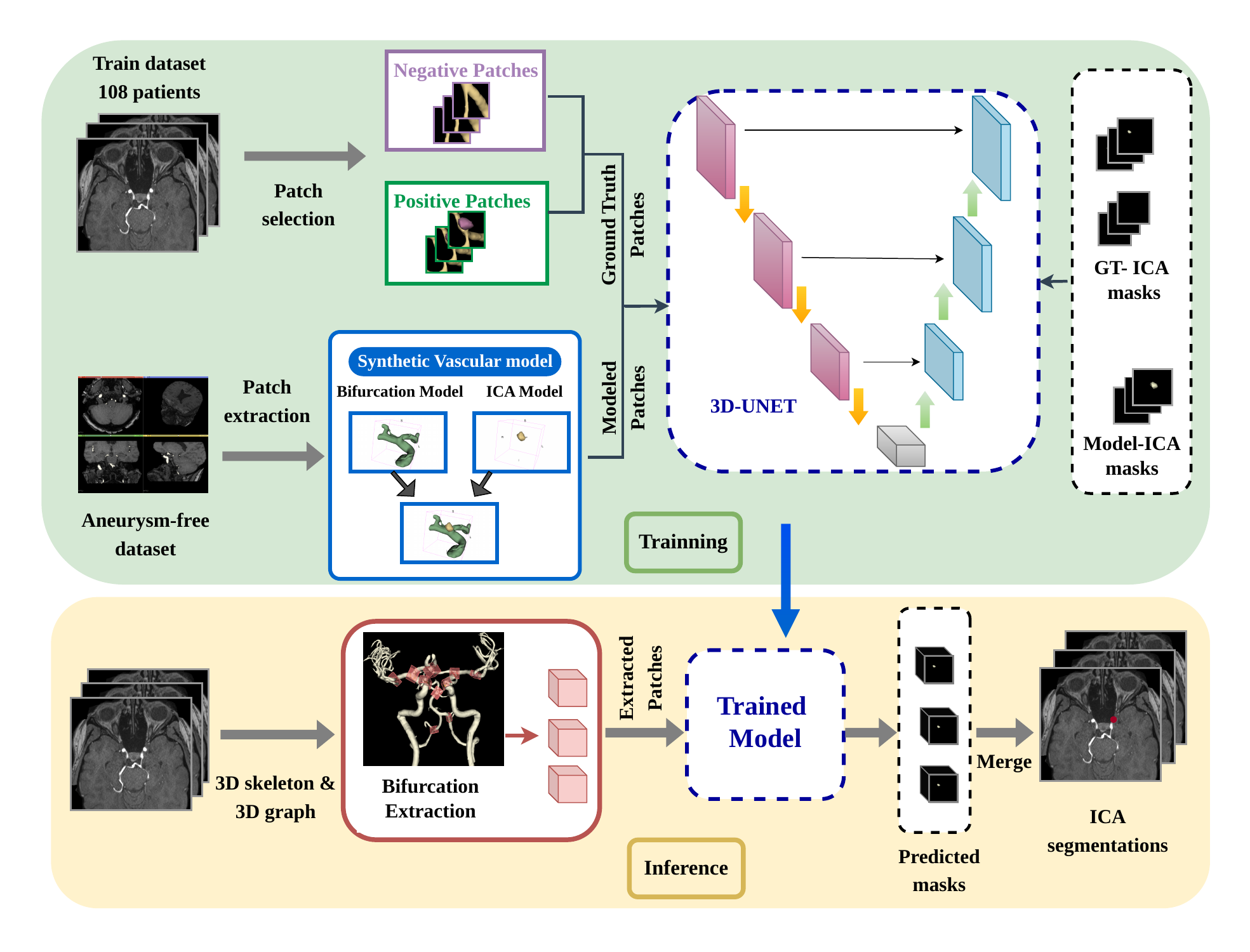}
\par\end{centering}
\caption{Overview of the global procedure, encompassing the training step using the synthetic images and the inference step. \label{fig:GlobalProcess}}
\end{figure*}

The main purpose of this synthetic model is actually to be able to effortlessly build up significant images datasets in order to efficiently train convolutional neural networks for pattern recognition tasks.
In this section, we present experimental results related to the use of synthetic images as a source of data augmentation within an intracranial aneurysm detection scenario. 
Indeed the synthetic model is used to generate a substantial training dataset to feed a 3D U-Net model.
Fig.~\ref{fig:GlobalProcess} shows an overview of the entire aneurysm detection process. The U-Net is trained on both the synthetically generated images and some actual TOF acquired from patients with aneurysm\footnote{The source code of the neural network (along with its weights) has been made available here: \url{https://gitlab.univ-nantes.fr/autrusseau-f/ica-detection}}. The trained U-Net model is then applied onto some extracted bifurcations during the inference phase.

We will first introduce the Deep Learning based  method being used, along with the corresponding dataset, the training strategy and the evaluation approach. Finally, we will evaluate the efficiency of our approach, and estimate the benefit of adjoining synthetic images for the CNN training step.

\subsection {DL based detection of intracranial aneurysms}

\subsubsection{Dataset}
For this study, a total of 190 TOFs scans of unruptured intracranial aneurysms were collected from over thirty different French institutions\footnote{Our study was conducted on retrospective and fully anonymized data; According to the french law, written consent was waved and our protocol was approved by local ethical group (GNEDS)}. 
Furthermore, 14 additional TOF images not containing any aneurysm were included in the test set to evaluate the performances of the developed model. 
Out of these 190 subjects, 58 had more than one aneurysm.
Overall 254 aneurysms are included in this study (presenting a mean radius of $2.49\pm0.82$ mm).
Tables \ref{tab2} and \ref{tab3} show their respective sizes and locations.

\begin{table}[!ht]
\caption{Aneurysms radii in the training and test data sets.\label{tab2}}
\centering{}%
\begin{tabular}{|c|c|c|c|}
\hline 
\textbf{ICA radius} & $\leq2$ mm & 2-3 mm & $>3$ mm\tabularnewline
\hline 
\textbf{Train} & 19 & 64 & 44\tabularnewline
\hline 
\textbf{Test} & 47 & 69 & 11\tabularnewline
\hline 
\end{tabular}
\end{table}

From now on, when evaluating the detection performances per bifurcation labels, we will group various labels altogether. Indeed, we have grouped the three bifurcations along the basilar artery together (M, N and O), as overall, few aneurysms pop out along the anterior portion of the Circle of Willis. Moreover, due to the anatomical configuration of the CoW on humans, both the PCoA and MCA arteries commonly reach the Internal Carotid Artery in the same vicinity, and hence, the bifurcations E (\textit{resp.} F) are very close to the bifurcations I (\textit{resp.} J), and thus, the aneurysms emerge in a very close neighborhood (possibly within the same cropped area encompassing two bifurcations of interest).

\begin{table}[!ht]
\caption{Distribution of the aneurysms based on their location within the CoW. \label{tab3}}
\centering{}%
\begin{tabular}{|c|c|c|c|c|c|c|c|}
\hline 
\textbf{Bifurcation} & A-B & C-D & E-F-I-J & G-H & K-L & M-N-O & Total\tabularnewline
\hline 
\textbf{Training set} & 11 & 1 & 12 & 54 & 48 & 1 & 127\tabularnewline
\hline 
\textbf{Test set} & 17 & 6 & 18 & 44 & 37 & 5 & 127\tabularnewline
\hline 
\end{tabular}
\end{table}

A trained operator (author F.A.) performed the annotations to build up the dataset. Subsequently, a neuroradiologist with 10 years of experience (author R.B.), carefully reviewed the cases to ensure the exclusion of any potential false positives or false negatives that might have been initially reported in the original annotation.

\subsubsection{Real patches selection and Neural network}

In our study, we have used a patch-based approach for the aneurysm detection process. Instead of using entire volumes, we feed the neural network with 3D patches of size $64^3$ voxels ($25,6^3$ mm). 
In order to select the 3D training patches, we employed a random extraction strategy. Specifically, for each aneurysm, we randomly extract 10 patches centered around the vicinity of the aneurysm (random shifts along the $x$, $y$ and $z$ directions). However, for each extracted patch, we ensure that the entirety of the ICA is included within the cropped area.
This approach ensures that the training dataset contains diverse samples representing different locations of the aneurysm within the extracted patches.
For negative samples, we extract some 3D patches encompassing some cerebral arteries, but without any aneurysm. For each  TOF volume, we extract 20 such patches.

We have opted for a segmentation network for the dual tasks of \textit{i)} ICA mask segmentation and \textit{ii)} the subsequent ICA detection.
The segmentation process uses a 3D U-Net architecture \cite{cicek20163d}, which follows an encoder/decoder structure. 

Each level in the encoder comprises two consecutive 3D convolution layers with a kernel size of $3\times 3 \times 3$ and a stride of 1, followed by a batch normalization layer and a Rectified Linear Unit (ReLU) activation. This sequence is followed by a max pooling layer with a kernel size of $2\times 2 \times 2$ and a stride of 1. The depth of the feature maps doubles with each downsampling step, starting from 32 and increasing up to 256. In the decoder path, max pooling layers are replaced with up-sampling layers. The ReLU activation function was applied to all layers, except for the final layer, which was followed by a sigmoid function.
The model was trained for 50 epochs with a batch size of 16 using  Adam optimizer with a learning rate of 0.0001. The Combo loss function \cite{Taghanaki2019} which combines both the Dice coefficient and the binary Cross-entropy loss was used in conjunction with this optimization algorithm. After each epoch, we evaluated the model on the validation set, and the model with the lowest validation loss was saved.

\subsubsection{Data Augmentation}
Overall, 134 TOFs free of any aneurysm were collected from the previous work described in \cite{NADER2023102919}.
These scans were utilized for constructing 3D synthetic cropped portions mimicking the characteristics of an original TOF, as explained in the previous section. For this purpose, a total of 998 synthetic patches were modeled. Each patch was specifically centered on the bifurcations of interest within the CoW.
Once the bifurcation has been accurately modeled, a synthetic aneurysm is incorporated thanks to the 3D model. This is achieved by varying the radius parameter and by applying degrees of elastic deformations (as explained in Sec.~\ref{ssec:ICApos}) to simulate the diverse characteristics observed in actual aneurysms (average modeled aneurysms radius: $2.33\pm0.52$).
Tables \ref{tab4} and \ref{tab5} show the distribution of the modeled aneurysms with respect to their locations and sizes.

\begin{table}[!ht]
\caption{Distribution of modeled patches for each bifurcation label. \label{tab4}}

\centering{}
\begin{tabular}{|c|c|c|c|c|c|c|c|}
\hline 
\textbf{Bifurcation} & A-B & C-D & E-F & G-H & I-J & K-L & M-N-O\tabularnewline
\hline 
\textbf{\# of ICAs} & 165 & 158 & 156 & 175 & 102 & 111 & 131\tabularnewline
\hline 
\end{tabular}
\end{table}

\begin{table}[!ht]
\caption{Number of modeled aneurysms per radius range. \label{tab5}}

\centering{}
\begin{tabular}{|c|c|c|c|}
\hline 
\textbf{Radius} & $\leq2$ mm &  $\in\, ]2, 3]$ mm & $>3$mm\tabularnewline
\hline 
\textbf{Count} & 292 & 596 & 110\tabularnewline
\hline 
\end{tabular}
\end{table}

\subsubsection{Evaluation approach}

\label{ssec:Eval}

To validate the possible improvements brought by the use of the synthetic model, we have conducted three separate experiments.
In the first experiment (\textit{Exp.\#1}), we have trained a baseline model using actual TOFs patches. To assess the performance of our model, we employed a four fold cross-validation approach. The dataset was split into four folds with each fold containing 27 samples.
During each cross-validation split, three folds (81 samples) were exploited to train the model, while the remaining fold (27 samples) was used for validation purposes and for hyperparameter optimization. This process was repeated four times, to ensure that all 108 TOFs were ultimately used for evaluation.
In the second experiment (\textit{Exp.\#2}), we trained another segmentation network but augmenting the training dataset with 998 synthetic patches. Similarly, we evaluated the performance of the model using the same cross-validation split. 
In \textit{Exp.\#3}, the training dataset of actual TOF patches was augmented using  traditional data augmentation operations,  namely rotations within the interval $[-15^\circ,+15^\circ]$ and $(90^\circ, 180^\circ, 270^\circ)$, as well as horizontal and vertical flipping.
Following the training phase, the model evaluation employs the holdout test set using the four-fold models. Subsequently, the resulting predictions from these four models were averaged to derive the final predictions.
In the inference phase, for the three mentioned experiments, we adopt a prior anatomical selection of patches. We only retain the patches being centered onto all the bifurcations of the cerebral vasculature.
By focusing on patches centered around the cerebral artery bifurcations, the inference process aims to target the regions being most susceptible to witness an aneurysm development. This approach is based  on the anatomical knowledge of aneurysm occurrence, enhancing the accuracy of the results. Hence, as previously explained, to select the corresponding patches, an automated vessel segmentation step was performed using a pre-trained U-Net segmentation algorithm.
The details of the specific pre-trained  network can be found in \cite{NADER2023102919}.
Then, a 3D undirected graph \cite{Nouri2020} is generated from the extracted skeleton to extract the corresponding bifurcations.

For all the experiments, we have analyzed the lesion-level sensitivity and the false positive rate (number of false positive per TOF). The lesion-level sensitivity quantifies the proportion of true aneurysms correctly identified by our method, while the false positive count per TOF provides insights about  the method  possibly producing an excessive number of false identifications, which could lead to an unnecessary burden for the neuroradiologists. To compute the detection performance evaluation metrics, we consider each segmented connected component issued from the binary output of the U-Net as a potentially detected object. Each connected component whose center of mass falls no farther away from the maximum radius of a true aneurysm mask, is thus considered as a true positive detection. Otherwise, it is regarded as a false positive detection.

In addition, we applied the evaluation protocol for assessing the segmentation performance of true (ICAs) as described in the ADAM challenge \cite{Timmins2021}. This protocol focuses on evaluating the segmentation metrics only for the true detected ICAs, excluding any false positives, to simulate how the tool could be practically used by neuroradiologists.

\subsection{Performance analysis}

Let us now examine the performances of our proposed method on the test set, in terms of global detection rate as well as on a per-bifurcation scenario. 

\subsubsection{Overall detection  performance}
As previously mentioned, the study compares the performances of three CNN training approaches. In \textit{Exp.\#1}, the CNN which is exclusively trained onto real data, successfully identified a total of 96 aneurysms within a dataset containing 127 instances, which corresponds to a lesion-level sensitivity of $75.60\%$.  In contrast, \textit{Exp.\#2} involves training the same CNN using a combination of real data patches and VaMos data patches. 
Notably, this sensitivity further improved to $88.97\%$ for \textit{Exp.\#2} with 113 detected aneurysms.
Meanwhile, when considering traditional data augmentation techniques in \textit{Exp.\#3}, the CNN identifies 104 aneurysms which corresponds to a lesion-level sensitivity of $81.88 \%$.

\subsubsection{False detections}
These results show a high diagnosis performance, with lesion-wise sensitivity notably improving by incorporating synthetic patches, reaching $89\%$ on the  test set. The synthetic data proved useful as a complementary tool to reduce the missed aneurysms rate. However, obtaining a high sensitivity may unfortunately lead to a slightly higher false-positive detection rate.
Overall, the network exhibits an average false positive rate of $0.22$ in \textit{Exp.\#1}, whereas  when incorporating synthetic patches, the false positive rate reaches $0.40$. Similarly, traditional data augmentation techniques employed in \textit{Exp.\#3} were associated with an increase of FP rate ($0.36$). 
The  increase in the false positive rate upon integrating synthetic patches, along with the significant  sensitivity gain, emphasizes the importance of adopting \textit{Exp.\#2}. This illustrates the trade-off between sensitivity and false positive rate and highlights the added value of the synthetic data augmentation.

Based on a thorough visual inspection of the false-positive detections, we have identified various reasons for these FP, including \textit{1)} The complex anatomy of the internal carotid artery, with sometimes rather strong variations in vessel diameter, a high tortuosity, and a significant bending, right below the ophtalmic artery that can be confused with a large aneurysm ;
\textit{2)} Brain arteries are susceptible to flow-related changes due to factors such as atherosclerosis (calcified plaque) or stenosis. These conditions can alter blood flow patterns and vessel appearance, potentially leading aneurysm-like vessels shapes, and hence, false-positives.
\textit{3)} At the emergence of a daughter artery on a bifurcation, at the very basis of the artery, an outpouching can be formed. In other words, the artery stars with a conic shape, exhibiting a broad base located at the bifurcation. This is clinically referred to as an \textit{infundibulum}~\cite{Infundibulum98}. Sometimes, such uncommon shapes can be mistakenly detected as being an aneurysm.
\textit{4)} Our vascular model has been designed in such a way to model thrombosed aneurysms. Indeed, we have noticed that for the larger circular aneurysms, a thrombosis often appears nearby the ICA center; the blood flow may circulate along the aneurysm walls, forming some sort of vortex, and hence, leading to a slower blood displacement toward the center (inducing a more radio-opaque area). Unfortunately, such a phenomenon can also occur along the Internal Carotid Artery, thus leading to false positive detections.

\subsubsection{Impact of aneurysms size and locations}
The performance variability when considering aneurysm size, is depicted in Table \ref{tab11}. \textit{Exp.\#1} yielded a detection rate of $0.5106$ for very small aneurysms (less than 2 mm), while \textit{Exp.\#2} achieved an improved rate of $0.7659$ and \textit{Exp.\#3} exhibited a detection rate of $0.6382$
The CNN was able to detect a greater proportion of aneurysms falling  within the size range of 2 mm to 3 mm. This trend was observed in the three experiments, with a notable increase observed for \textit{Exp.\#2} ($0.8840$ and $0.9130$ versus $0.9565$). All methodologies, however, demonstrated a  detection rate of 1 for aneurysms with a radius exceeding 3 mm.

\begin{table}[!ht]
\caption{Lesion-level sensitivity according to the aneurysm size for the test
set \label{tab11}}

\centering{}%
\begin{tabular}{|c|c|c|c|}
\hline 
\textbf{Radius} & $\leq2$ mm & $\in\, ]2, 3]$ mm & $>3$ mm\tabularnewline
\hline 
\textbf{\textit{Exp.\#1}} & 51.06 $\%$ & 88.40 $\%$ & 100 $\%$\tabularnewline
\hline 
\textbf{\textit{Exp.\#2}} & 76.59 $\%$ & 95.65 $\%$ & 100 $\%$\tabularnewline
\hline 
\textbf{\textit{Exp.\#3}} & 63.82 $\%$ & 91.30 $\%$ & 100 $\%$\tabularnewline
\hline 
\end{tabular}
\end{table}

\begin{figure}[!ht]
\begin{centering}
\includegraphics[width=0.8\columnwidth]{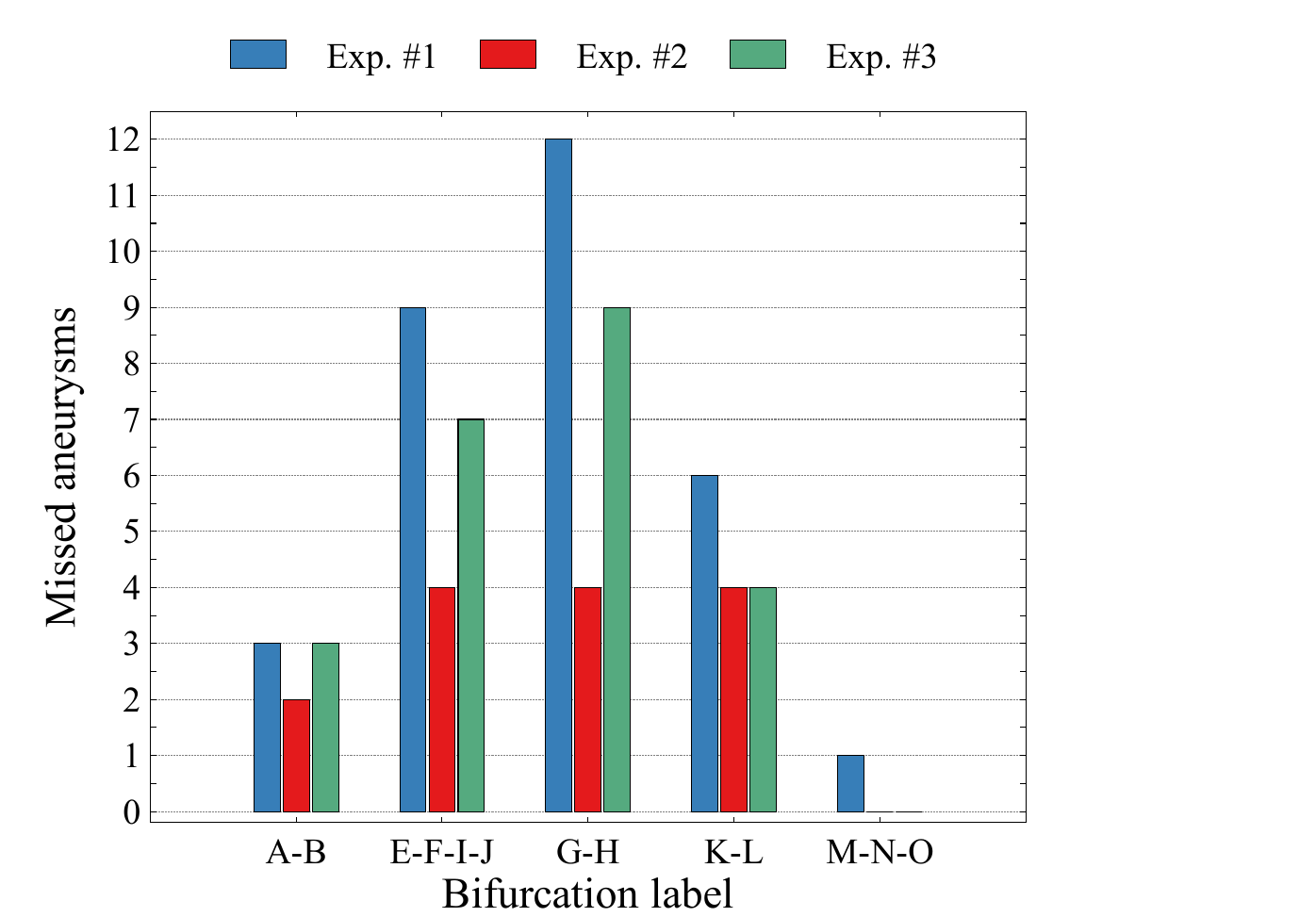}
\par\end{centering}
\caption{Missed detections with respect to the aneurysms positions (along the Circle of Willis) in the test dataset for all three tested experiments.\label{fig10}}
\end{figure}

Regarding the impact of aneurysm location, Fig.~\ref{fig10} depicts the number of missed aneurysms per bifurcation of interest. Specifically, in \textit{Exp.\#1}, the CNN missed a higher number of aneurysms located along the Middle Cerebral Artery (MCA) (G-H bifurcations), as well as along the Internal Carotid Artery, with a significant concentration in the bifurcation segment separating this artery into the smaller Ophthalmic Artery (OA) (labels K \& L). Additionally, missed aneurysms are observed within the branches of the Anterior Cerebral Artery (ACA) and the Posterior Communicating Artery (PCOM), which corresponds to bifurcations (E, F, I \& J).

In contrast, for \textit{Exp.\#2}, there is a substantial decrease in the number of missed aneurysms within the three locations cited  below.
The impact of the aneurysm location on the detection rates for the three experiments is presented in Table \ref{tab12}.
One can notice on this Table that for each bifurcation label, the amount of collected aneurysms matches remarkably well the percentages of occurrences as previously described on Fig.~\ref{fig:SchematicWillis}.
Hence anatomically, we are able to collect fewer aneurysms onto certain bifurcations.
Moreover, it is important to highlight that fewer synthetic aneurysms can be modeled onto the bifurcations labeled C, D, N and O (\textit{c.f.}, Table~\ref{tab4}) as, quite often, during the angiography exam, these bifurcations (at the farther ends of the MRA 3D stack) may be cropped out of the acquisition area.

\begin{table*}[!ht]
{\small{}\caption{The detection rates with respect to aneurysms location \label{tab12}}
}{\small\par}
\centering{}%
\begin{tabular}{|c|c|c|c|c|c|c|c|}
\hline 
\multicolumn{2}{|c|}{\textbf{Label}} & \textbf{A-B} & \textbf{C-D} & \textbf{E-F-I-J} & \textbf{G-H} & \textbf{K-L} & \textbf{M-N-O}\tabularnewline
\hline 
\multicolumn{2}{|c|}{\textbf{Count}} & 17 & 6 & 18 & 44 & 37 & 5\tabularnewline
\hline 
\multirow{3}{*}{\textbf{Detection (\%)}} & \textbf{Exp. \#1} & 82.35 & 100.00 & 50.00 & 76.19 & 83.78 & 80.00\tabularnewline
\cline{2-8} \cline{3-8} \cline{4-8} \cline{5-8} \cline{6-8} \cline{7-8} \cline{8-8} 
& \textbf{Exp. \#2} & 88.23 & 100.00 & 77.77 & 90.90 & 89.18 & 100.00\tabularnewline
\cline{2-8} \cline{3-8} \cline{4-8} \cline{5-8} \cline{6-8} \cline{7-8} \cline{8-8} 
& \textbf{Exp. \#3} & 82.35 & 100.00 & 61.11 & 79.54 & 89.18 & 100.00\tabularnewline
\hline 
\end{tabular}
\end{table*}

\subsubsection{Segmentation performance of true ICAs}

\begin{figure}[!ht]
\centering
\includegraphics[width=0.56\columnwidth]{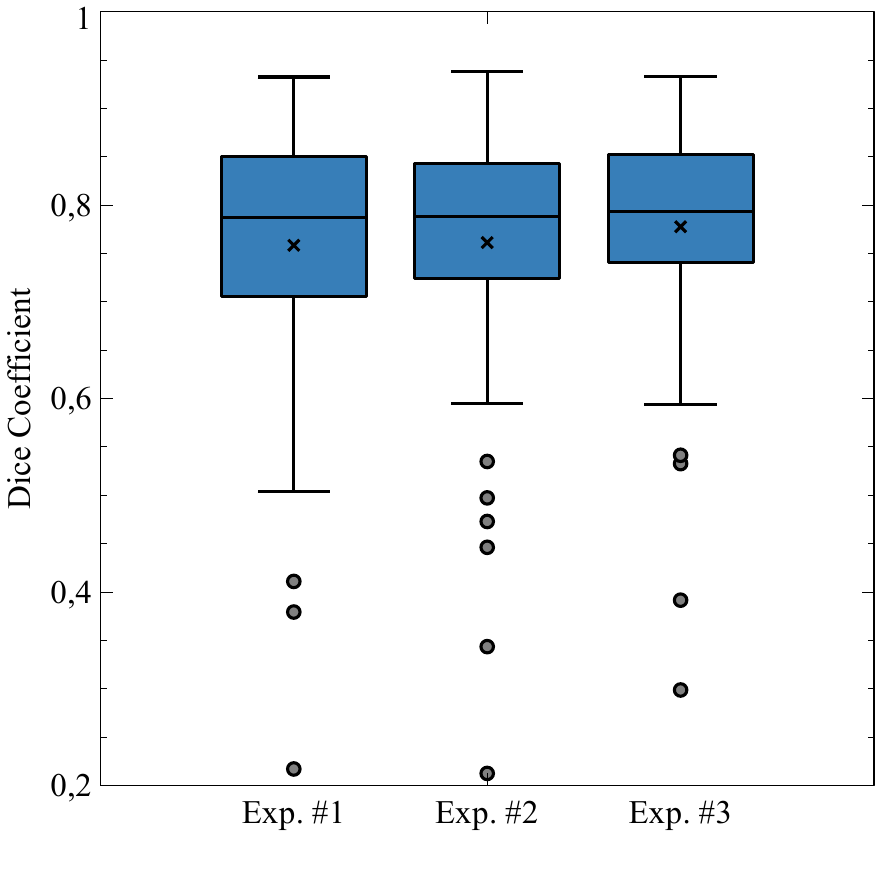}
\caption{\label{fig14} Dice similarity coefficient of true ICAs for all three tested experiments.}
\end{figure}

Fig.~\ref{fig14} displays the distribution of Dice score for the detected aneurysms within the three experiments $\#1$,  $\#2$ and $\#3$.  The average Dice score for \textit{Exp.\#1}  is 0.7585 $(\pm  0.13$) which indicates a fairly good similarity with the actual aneurysms masks. \textit{Exp.\#2} and \textit{Exp.\#3} achieve a  comparable Dice score of 0.7613 ($\pm 0.12$) and 0.77 ($\pm 0.11$). It is important to note that a direct comparison between the three Dice coefficients is to be considered carefully, as the number of detected aneurysms differs between the  experiments.

However, the visual analysis of the box plots (Fig.~\ref{fig14}) representing the segmentation performance for \textit{Exp.\#1}, \textit{Exp.\#2}  and \textit{Exp.\#3} reveals similarities in their overall appearance. 
\textit{Exp.\#2} and \textit{Exp.\#3} appear to have a slightly denser distribution of Dice coefficients.
Such outliers can be attributed to the smaller aneurysms.

\begin{table}
\centering
\caption{Ablation study exploring the impact of different VaMos noise distributions and increased number of augmented data on the detection performance.\label{tab:ablation}}
\begin{tabular}{|c|c|c|}
\hline
\textbf{Method} & \textbf{Sensitivity(\%)} & \textbf{FP/case}\tabularnewline
\hline 

\textbf{Noise Distribution} &  & \tabularnewline
\hline 
Gaussian &86 & 0.40\tabularnewline

Rician& 85& 0.50 \tabularnewline

Perlin & 81& 0.56 \tabularnewline
\hline

\textbf{Training data} &  & \tabularnewline
\hline
TOF & 76 & 0.22\tabularnewline

TOF+A1& 77 & 0.51\tabularnewline

TOF+A2 & 80& 0.46 \tabularnewline

TOF+A3 &89 & 0.40 \tabularnewline
\hline
\end{tabular}
\end{table}

\subsubsection{Ablation study}
To validate the various constituents of VaMos, we performed an ablation study on the aneurysm detection task, using different configurations. 

In order to assess the impact of the noise on the overall detection performances, we have modeled the same augmented ICA patches but with different noise distributions \cite{aja2016statistical}: Gaussian, Perlin and Rician. The results of our experiments are summarized in Table \ref{tab:ablation}.
Although Gaussian and Rician  displayed similar effectiveness, Gaussian noise demonstrated a marginally higher sensitivity and a slightly better false positive detections. Perlin noise, on the other hand, demonstrated lower sensitivity (81\%) and a higher false positive rate (0.56 per case).
Given these factors, Gaussian noise is deemed more advantageous for our specific task of data augmentation.

Furthermore, we compared the performance of the detection network using increased number of (VaMos) augmented patches size : A1 (334 patches), A2 (622 patches) and A3 (998 patches).
Obviously, for all three configurations,  the test was conducted on the holdout test data. As illustrated in Table \ref{tab:ablation}, the addition of synthetic patches consistently improves sensitivity. This suggests that the model becomes better at identifying true positives with more diverse training data.
The initial spike in false positives with (TOF+A1) indicates that while synthetic patches help in identifying more true positives, they also introduce noise, leading to more false positives. However, with further augmentation (TOF+A2 and TOF+A3), the model seems to learn to better distinguish between true positives and false positives, as witnessed by the reduction in FP/case.

\subsubsection{Comparison with nnU-Net}
We have assessed the effectiveness of our method on the traditional detection task by comparing it with the fully automated state-of-the-art baseline, nnU-Net \cite{isensee2021nnu}. nnU-Net is a self-configuring deep learning framework designed for biomedical image segmentation that automatically adapts its architecture and hyperparameters to optimize performance for different datasets. To ensure a fair comparison and reproducibility of the results, we employed 4-fold cross-validation. Our method showed competitive performance  achieving a sensitivity of $89\%$ associated with a FP/case of 0.4. In comparison, nnU-Net achieves a  sensitivity of $82\%$ with a lower FP/case of 0.125. Table \ref{tab17} illustrates the reason for our higher sensitivity, which is attributed to higher detection of small aneurysms (15 $\%$ increase). VaMos introduces variability in small aneurysms. Such a shape variability cannot be produced by traditional augmentation, and hence cannot be considered within nnU-Net. 

\begin{table}[!ht]
\caption{Comparison of the sensitivity between our method  and nnU-Net with repesct to aneurysms size \label{tab17}}
\centering{}%
\begin{tabular}{|c|c|c|c|}
\hline 
\textbf{Radius} & $\leq2$ mm & $\in\, ]2, 3]$ mm & $>3$ \tabularnewline
\hline 
\textbf{\textit{nnU-Net}} & 61.7 $\%$ & 92.75 $\%$ & 100 $\%$\tabularnewline
\hline 
\textbf{\textit{OURS}} & 76.59 $\%$ & 95.65 $\%$ & 100 $\%$\tabularnewline
\hline 
\end{tabular}
\end{table}

\section{Discussion and Conclusion}
\label{sec:Discussion}

In this section, we analyze the contribution brought by the synthetic vasculature model, being able to faithfully counterfeit (while cleverly altering) portions of TOF images.
Indeed, the synthetic model is composed of various processes, including a meticulous modeling of the cerebral arteries and bifurcations geometry, alongside the introduction of surrounding noise and finally, embedding aneurysms of various sizes and shapes.
Our goal is to provide a substantial dataset that may improve the performances of several deep learning tasks including the segmentation or detection of the cerebral aneurysms.
A salient highlight of our work is the successful generation of synthetic aneurysms with varying sizes, shapes, and locations. These artificial aneurysmal sacs have been integrated into modeled MRA scans originally lacking any aneurysm, thus resulting in an augmented dataset that aligns more closely with real-world scenarios. The important focal point of this approach is the strategic positioning of cerebral aneurysms within the Circle of Willis, enhancing the fidelity of the simulated model.

As the main finding of this study, the CNN trained using a combination of both genuine and synthetic patches led to a significantly improved sensitivity in detecting intracranial aneurysms compared to a CNN trained solely on TOF data and/or augmented with traditional data augmentation techniques. While the network missed $24.4\%$ of the lesions on the test data in \textit{Exp.\#1}, including the VaMos  patches during the training step significantly improved the ICA detection performances of the CNN. Indeed, the network  missed a smaller portion of aneurysms (only $11\%$) on the test set. Only 14 aneurysms were missed in \textit{Exp.\#2}: eleven were tiny aneurysms, two exhibited uncommon shapes (high order spherical aberrations), and finally, one was presenting a shape strongly similar to an infundibulum (no clearly delineated aneurysm neck).
Traditional data augmentation techniques  led to a modest increase in sensitivity ($4\%$) compared to the significant improvement seen with VaMos ($13\%$). This disparity highlights the limitations of traditional methods. These conventional techniques mainly improve the model's ability to generalize from existing data but fail to introduce substantial diversity in critical anatomical features like aneurysm shape and size. Consequently, they are less effective in training models to identify a wide variety of aneurysms across different patients. On the other hand, advanced generation models like VaMos generate more complex and varied synthetic examples, incorporating significant variations in aneurysm characteristics, which enhances the model's performance.

This research has successfully achieved a high level of detection sensitivity, showcasing its potential as supplementary tool for neuroradiologists to address the issue of overlooked aneurysms.
Nevertheless, it is important to note that the integration of synthetic patches, while effective in boosting sensitivity, can potentially contribute to a slightly increased false-positive rate.
This issue might arise if the synthetic patches actually introduce irregular aneurysms shapes that do not very faithfully reflect the genuine aneurysm sacs as acquired on TOFs. However, the training methodology, employed alongside with a judicious selection of the synthetic model parameters has yielded an improved  detection performance while maintaining an acceptable rate of false-positive detections. Specifically, the false-positive detection rate stood at 0.40 within \textit{Exp.\#2}, in contrast to 0.22 in \textit{Exp.\#1}. While the increase in the FP from 0.22 to 0.4 represents an $18\%$ absolute increase, this is often considered reasonable within the healthcare context. False positives primarily lead to additional diagnostic procedures, which, although not ideal, are less harmful compared to the risks associated with undetected aneurysms.
The ability to maintain a minimal count of false-positive detection rate can also be explained by 2 factors:
To begin with, the use of a prior selection of patches (3D undirected graph generated from the skeleton, as explained in sec.~\ref{ssec:Eval}) during the inference phase by extracting patches around vascular bifurcations. Consequently, this approach minimizes the susceptibility of the algorithm to confuse non-vessel structures and reduces the likelihood of incorrect predictions.
Furthermore, the final prediction on the test set is derived by aggregating the probabilities obtained from 4 cross validation models. This strategy enhances the overall robustness of the model's predictions.
Moreover, the mean Dice score index of true ICAs is 0.76 which is relatively high. This suggests that this automatic segmentation method performs at a similar level to manual segmentation once the true ICA has been correctly identified. The automatic segmentation could save time and effort in the analysis of medical imaging data and potentially improve the efficiency in diagnosing and analyzing ICAs.

Regarding the aneurysm size, our analysis demonstrated that there were no significant discrepancy in sensitivity for large aneurysms. Indeed, the sensitivity reached $100\%$ for aneurysms having a radius larger than 3 mm for all experiments.
The CNN achieved an overall sensitivity of $51.06\%$ for detecting aneurysms being smaller than 2 mm in \textit{Exp.\#1}, which complies with common findings in state-of-the-art aneurysm detection studies. For such small aneurysms, the sensitivity may increase with the number of training cases involved. Notably, in \textit{Exp.\#2}, the sensitivity value increased to $76.59\% $ by training the model with small sizes synthetic aneurysms. Nevertheless, enhancing the diagnostic performance of the model in detecting small aneurysms may need further exploration and training with a larger set of synthetic cases representing small aneurysms.

With regard to the sensitivity at different locations, the \textit{Exp.\#2} achieved better performances than \textit{Exp.\#3} for aneurysms situated onto the bifurcations G-H, E-F-I-J. It is important to note that this sensitivity increase appears to be more strongly linked to the size of aneurysms in these specific locations rather than solely on the location itself. In fact, the detection rates stratified according to aneurysm location and size were compared using  Fishers’ exact test. No significant difference was found between different locations (p-value=0.69).
However, the statistical analysis revealed a significant difference in aneurysm detection rates when stratified by different sizes (p-value=0.005).

Indeed, it is important to interpret the sensitivity values for the categories C-D and M-N-O  with careful consideration. Particularly so, as the dataset comprises only few aneurysms in those locations compared to other locations. For future investigations, it could be valuable to explore the CNN efficiency with a larger population, \textit{i.e.}, incorporating more aneurysms located in regions such as M-N-O, and C-D.

This study has been conducted on TOF acquisitions only, however, the model is, in its very nature, quite flexible and its adaptation to other modalities, such as CTA or DSA, should be relatively straightforward. We have made an attempt to model DSA acquisitions, very slight modifications were brought to the model (the fluids are not visible within DSA acquisitions, and hence the corresponding background noise had to be removed). Eventually, the modeled DSA patches were reasonably well modeled. Interested reader may refer to the GitLab repository for some examples.  
The operating mode remains the same, only the background noise modeling might need some slight adjustments.

\bibliographystyle{IEEEtran}

\end{document}